\documentclass[runningheads]{llncs}
\usepackage{wrapfig}

\usepackage{eccv}

\usepackage{eccvabbrv}

\definecolor{forestgreen}{RGB}{34,139,34}
\definecolor{alignment}{RGB}{0,173,103}
\definecolor{misalignment}{RGB}{237,125,49}
\definecolor{unalignable}{RGB}{200,0,2}

\newcommand{\myparagraph}[1]{\vspace{3pt}\noindent\textbf{#1.}}
\definecolor{stage1}{RGB}{8,90,214}
\definecolor{stage2}{RGB}{32,156,56}
\definecolor{alignable}{RGB}{120,158,1}
\definecolor{unalignable}{RGB}{253,191,110}
\newcommand{\modelname}{\textbf{NaSVA}}
\usepackage{pifont}
\newcommand{\cmark}{\ding{51}}
\newcommand{\xmark}{\ding{55}}
\usepackage{graphicx}
\usepackage{booktabs}
\usepackage{multirow}
\usepackage{microtype}
\usepackage{makecell}
\usepackage{caption}
\usepackage{wrapfig}

\usepackage[accsupp]{axessibility}  

\usepackage{hyperref}

\usepackage{orcidlink}

\begin{document}

\title{Multi-Sentence Grounding for Long-term Instructional Video} 


\author{Zeqian Li\inst{1\star}\orcidlink{0000-0002-9721-8668} \and
Qirui Chen\inst{1\star}\orcidlink{0009-0002-0330-1688} \and
Tengda Han\inst{3}\orcidlink{0000-0002-1874-9664} \and 
Ya Zhang\inst{1,2}\orcidlink{0000-0002-5390-9053} \and 
\\
Yanfeng Wang\inst{1,2}\orcidlink{0000-0002-3196-2347} \and
Weidi Xie\inst{1,2,3}\orcidlink{0009-0002-8609-6826}}

\authorrunning{Z. Li et al.}

\institute{Coop. Medianet Innovation Center, Shanghai Jiao Tong University \and
Shanghai AI Laboratory, China \and
Visual Geometry Group, University of Oxford \\
\url{https://lzq5.github.io/Video-Text-Alignment/}}

\maketitle

\renewcommand{\thefootnote}{\fnsymbol{footnote}}
\footnotetext[1]{These authors contributed equally to this work.}





\begin{abstract}
In this paper, we aim to establish an automatic, scalable pipeline for denoising the large-scale instructional dataset and construct a high-quality video-text dataset with multiple descriptive steps supervision, named \textbf{HowToStep}. 
We make the following contributions:
(i) improving the quality of sentences in dataset by upgrading ASR systems to reduce errors from speech recognition and prompting a large language model to transform noisy ASR transcripts into descriptive steps;
(ii) proposing a Transformer-based architecture with all texts as queries, iteratively attending to the visual features, to temporally align the generated steps to corresponding video segments.
To measure the quality of our curated datasets, we train models for the task of multi-sentence grounding on it, {\em i.e.}, given a long-form video, and associated multiple sentences, to determine their corresponding timestamps in the video simultaneously,
as a result, the model shows superior performance on a series of multi-sentence grounding tasks, surpassing existing state-of-the-art methods by a significant margin on three public benchmarks, namely, 9.0\% on HT-Step, 5.1\% on HTM-Align and 1.9\% on CrossTask. 
All codes, models, and the resulting dataset have been publicly released.

\keywords{ Instructional Video Understanding \and Video-Text Dataset \and Multi-Sentence Grounding}
\end{abstract}
   
\section{Introduction}
\label{sec:intro}

The research on visual-language representation learning has recently made great progress.
It is primarily driven by contrastive learning on image-caption pairs crawled from the Internet at scale~\cite{radford2021learning, li2022blip, li2023blip}, and has shown remarkable performance on zero-shot image classification. However, in videos, the time dimension adds extra complexity that requires temporally corresponding captions/descriptions, posing challenges to learning fine-grained representation for video understanding tasks, such as temporal action localization, visual-language grounding, and grounded visual question answering.


\begin{figure*}[!t]
  \centering
   \includegraphics[width=0.9\linewidth]{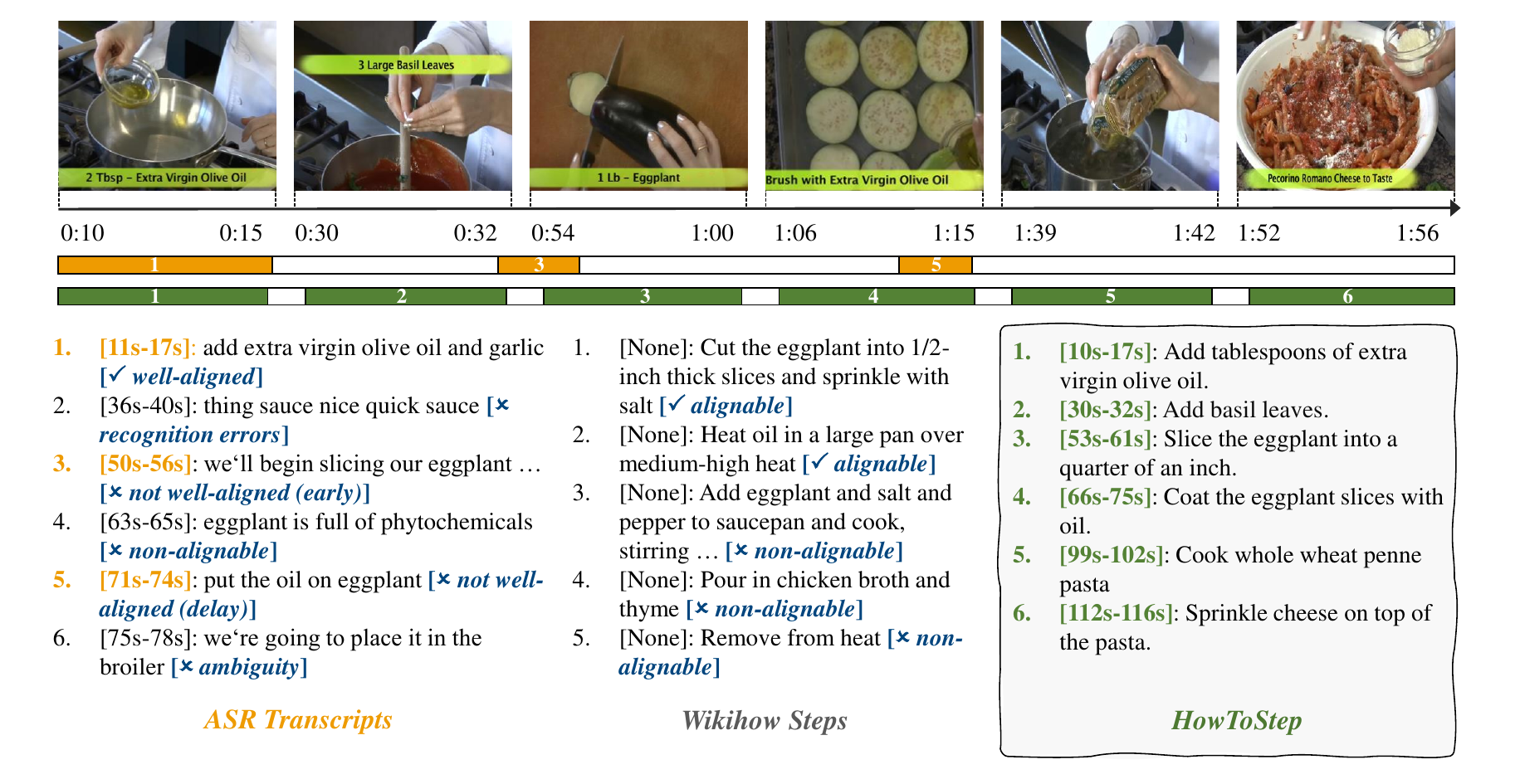}
   \caption{\textcolor{black}{A comparison of the proposed \textbf{HowToStep} with annotations on HowTo100M. Our dataset consists of multiple descriptive steps, with the corresponding temporal windows. Compared to existing training data derived from ASR transcripts~\cite{han2022temporal} and task-related articles of Wikihow~\cite{afouras2024ht,mavroudi2023learning,chen2022weakly}, 
   HowToStep data offers the following advantages: \\  \emph{1) Descriptive}: clearly describes the procedural action steps in the instructional video; \\  \emph{2) Concise}: all sentences can be grounded in the video, without redundancy or noises; \\ \emph{3) Temporally well-aligned}: offers precise temporal boundaries for procedural steps.}}
   \label{fig:example}
\end{figure*}

Instructional videos, {\em e.g.}, HowTo100M~\cite{miech2019howto100m}, have been widely used for learning video representations~\cite{mavroudi2023learning,lin2022learning,han2022temporal,miech2020end}, 
with the textual narrations acquired from the Automatic Speech Recognition~(ASR) system. 
However, unlike manually annotated captions, training models with these ASR transcripts naturally incur three issues: (i) off-the-shelf ASR systems may introduce recognition errors; 
(ii) spoken narrations are generally not of descriptive style, thus contain redundancy or ambiguity, {\em e.g.}, talking about a specific ingredient or using ambiguous pronouns (it, them, {\em etc.}); 
(iii) transcripts may not be well aligned with the visual signals, 
{\em e.g.}, greetings from the speaker, associating with inaccurate timestamps, or describing the action before/after performing it. 
According to the statistics from~\cite{han2022temporal}, 
only 30\% of the narrations are visually alignable, 
and only 15\% are naturally well-aligned with correct start/end timestamp, as presented in \cref{fig:example}.

In this paper, we aim to establish an automatic, scalable pipeline for `denoising' the large-scale instructional dataset, and contribute a `cleaned' video-text dataset, termed as \textbf{HowToStep}. 
Specifically, our curation procedure involves the following aspects, 
{\em first}, to improve the quality of text descriptions, we replace the original YouTube transcripts with the ones generated from WhisperX~\cite{bain2022whisperx}, 
and prompt the large language models (LLMs) to transform noisy ASR transcripts into coherent, descriptive steps that are closely related to the video content; 
{\em second}, to temporally align the texts to the corresponding video timestamp,
we adopt a two-stage determination procedure, {\em i.e.}, approximate estimation based on timestamps from ASR transcripts, followed by training a lightweight Transformer-based model for further refinement, termed as \textbf{Na}rrations / \textbf{S}teps to \textbf{V}ideo \textbf{A}ligner (\textbf{\modelname}), where we use multiple sentences as queries, to iteratively attend the video features, and output the alignability or optimal temporal windows.

As a result, to demonstrate the effectiveness of our dataset curation procedure, we evaluate the model trained on such dataset, namely \modelname, 
on multi-sentence grounding tasks - narrations alignment~\cite{han2022temporal} and procedural steps grounding~\cite{mavroudi2023learning}, both aiming to localize the corresponding temporal segments for multiple sentences in a video. Contrary to the existing text-to-video grounding tasks that focus on only a single sentence at a time, for example, temporal action localization~\cite{gao2017tall}, moment retrieval~\cite{lei2021detecting} and natural language queries~\cite{lin2022egocentric}, 
multi-sentence grounding necessitates the understanding of long instructional videos, with finer-grained events or actions, while managing multiple interrelated text queries simultaneously. 
Our model sets state-of-the-art performance on both procedural steps grounding and narrations alignment across three public benchmarks, surpassing existing models by a large margin, specifically, 9.0\% on HT-Step~\cite{mavroudi2023learning}, 5.1\% on HTM-Align~\cite{han2022temporal} and 1.9\% on CrossTask~\cite{zhukov2019cross}.

\section{Related Work}
\label{sec:related}


\myparagraph{Large-Scale Video-Text Datasets}
Multi-modal video datasets are crucial for video understanding tasks.
Conventional video-text datasets \cite{caba2015activitynet, zhou2018towards} are often manually labelled, suffer from short video lengths, limited scale, 
and coarse label granularity, and prevent the model from learning a generalized video representation. In the recent literature, scalability becomes an essential factor for constructing datasets, Youtube-8M~\cite{abu2016youtube} collects YouTube videos with metadata provided by the users,  Instagram65M \cite{ghadiyaram2019large} uses the associated hashtags as labels to supply weak supervision for training. 
In order to get descriptive sentences with richer semantics, instructional videos are collected at scale, as they naturally come with dense narrations, obtained from ASR systems, by far, the largest video-text dataset is HowTo100M \cite{miech2019howto100m}. 
As an alternative, \cite{Bain21} comprises over two million videos with weak captions scraped from the internet, while the captions are manually generated, they are not temporally aligned, and thus are insufficient for learning fine-grained temporal representation. 
In this work, we transform the noisy ASR transcripts into descriptive procedural steps and propose to train a model to improve the video-text correspondence for instructional videos, mitigating the flaws of the HowTo100M dataset for visual representation learning.




\myparagraph{Language Grounding in Videos}
Early efforts in language grounding predominantly concentrate on single sentence grounding tasks such as temporal action localization~\cite{gao2017tall}, moment retrieval~\cite{liu2018cross,zhang2019man,lei2021detecting}, and natural language queries~\cite{lin2022egocentric,ramakrishnan2023naq}, which lead to unprecedented progress in recent years. Given a temporally untrimmed video and a natural language query, the goal of single-sentence grounding is to determine the start and end times for the described content in the video. However, due to the limitation of query lengths to single sentences, the trained models lack an understanding of textual context.
Unlike single-sentence grounding, multi-sentence grounding involves simultaneously grounding multiple sentences from different semantic scales within the video. 
The model needs to determine the intervals corresponding to each sentence based on the correlation between multiple sentences, significantly increasing the task's complexity. This also enables the model to learn enhanced video-text representations.
Initially, \cite{bojanowski2014weakly, zhukov2019cross, dvornik2023stepformer} tries to delineate the video segments corresponding to an action list. Instead of an action list, scripts describing a series of events in the video are given for transcript alignment \cite{cour2008movie, pramod2009subtitle}.
The availability of large-scale video-text datasets such as HowTo100M has prompted many works on joint video-text embedding training.
Specifically, TAN \cite{han2022temporal} investigated directly aligning contextualized narration representations generated from ASR transcripts to video segments. 
Given that the ASR can be rather noisy, DistantSup~\cite{lin2022learning} proposes using distant supervision from a textual knowledge base, namely Wikihow~\cite{koupaee2018wikihow} to denoise narrations from ASR. 
More recently, instead of aligning narrations, 
VINA~\cite{mavroudi2023learning} proposes to ground procedural steps sourced from instructional article, 
however, as the order of instructions from Wikihow does not necessarily follow those in the video, grounding procedural steps is thus more challenging. 
In this paper, our improved HowToStep dataset enables to train models that tackle the two problems simultaneously, namely, narration alignment~\cite{han2022temporal} and procedural step grounding~\cite{mavroudi2023learning}, while previous studies have only focused on one aspect. 

\myparagraph{Dataset Curation with Large Language Models}
In the recent literature, large language models~(LLMs) such as GPT~\cite{radford2019language,brown2020language,neelakantan2022text} and Alpaca~\cite{taori2023alpaca} have achieved great success in natural language processing.
Constructing multi-modal datasets while generating pseudo-labels using LLMs becomes an efficient way to exploit the common sense knowledge in LLMs, and save human efforts for annotations.
For instance, VQA-T \cite{yang2021just} generates question-answer pairs for instructional video ASR transcripts by LLMs, which are then coupled with the related video. 
The VQA model pre-trained on the generated dataset exhibits enhanced generalization capabilities in various downstream tasks.
Similarly, LLMs are adopted to automatically derive question-answer pairs at scale for images or videos, by leveraging the existing image/video captions~\cite{changpinyo2022all,di24egoQA}. In addition to generating a dataset directly, some works use the LLMs to create pseudo-labels for large-scale video data that are later used for multi-modal vision tasks. 
For example, LAVILA~\cite{zhao2023learning} first trains a video captioning model on egocentric videos, and proposes to rephrase the densely generated captions from the model.
After finetuning, the model demonstrates improved performance.
As a concurrent work, \cite{shvetsova2023howtocaption} adopts LLMs to transform the ASR transcripts of instructional videos, however, there exists one crucial difference, we transform the transcripts into concise steps, rather than dense captions as in their work.

\section{Method}
\label{sec:method}

In this section, our goal is to establish an automatic, scalable pipeline for ‘denoising’ the large-scale instructional dataset, and contribute a textually descriptive, and temporally-aligned video-text dataset, which is termed as \textbf{HowToStep}. As presented in \cref{fig:pipeline}, the entire pipeline can be divided into three parts: 
(i) we leverage LLM to transform narrations from the ASR transcript into descriptive procedural steps; (ii) we use the similarity between the original transcript and generated steps to approximate the start/end timestamp for each step; 
(iii) we train a lightweight multi-sentence grounding model on the generated steps with the approximated timestamp, and then use the trained model to refine the time range for each generated step~(\ie, self-training). 

In the following sections, we start with improving the quality of texts by LLMs in \cref{subsec:llm}. Then, we introduce the multi-sentence grounding network that enables to align texts to their corresponding video segments in \cref{subsec:network}. Afterward, we describe the self-training procedure to refine the temporal ranges for the generated steps in \cref{subsec:self-training}. Lastly, we discuss two tasks that involve multi-sentence grounding, that can be addressed by our model, and reflect the quality of our curated dataset, namely, narration alignment and procedural step localization in \cref{subsec:app}.

\subsection{Noisy ASR Transcripts $\rightarrow$ Descriptive Procedural Steps}
\label{subsec:llm}

To improve the quality of texts of the ASR transcripts in HowTo100M, 
we first replace the original Youtube ASR transcripts with a recent WhisperX~\cite{bain2022whisperx} as the speech recognition tool,
which can potentially resolve errors in recognition or punctuation. 
However, some complex cases such as ambiguous pronouns and grammatical mistakes can not simply addressed by upgrading the ASR system. Therefore, we further propose to exploit the strong reasoning ability in large language models~(LLMs), to summarize the procedural action steps from the noisy ASR transcripts, converting them to descriptive textual explanations for instructional videos. The details are as follows:

\begin{figure*}[!t]
  \centering
   \includegraphics[width=\linewidth]{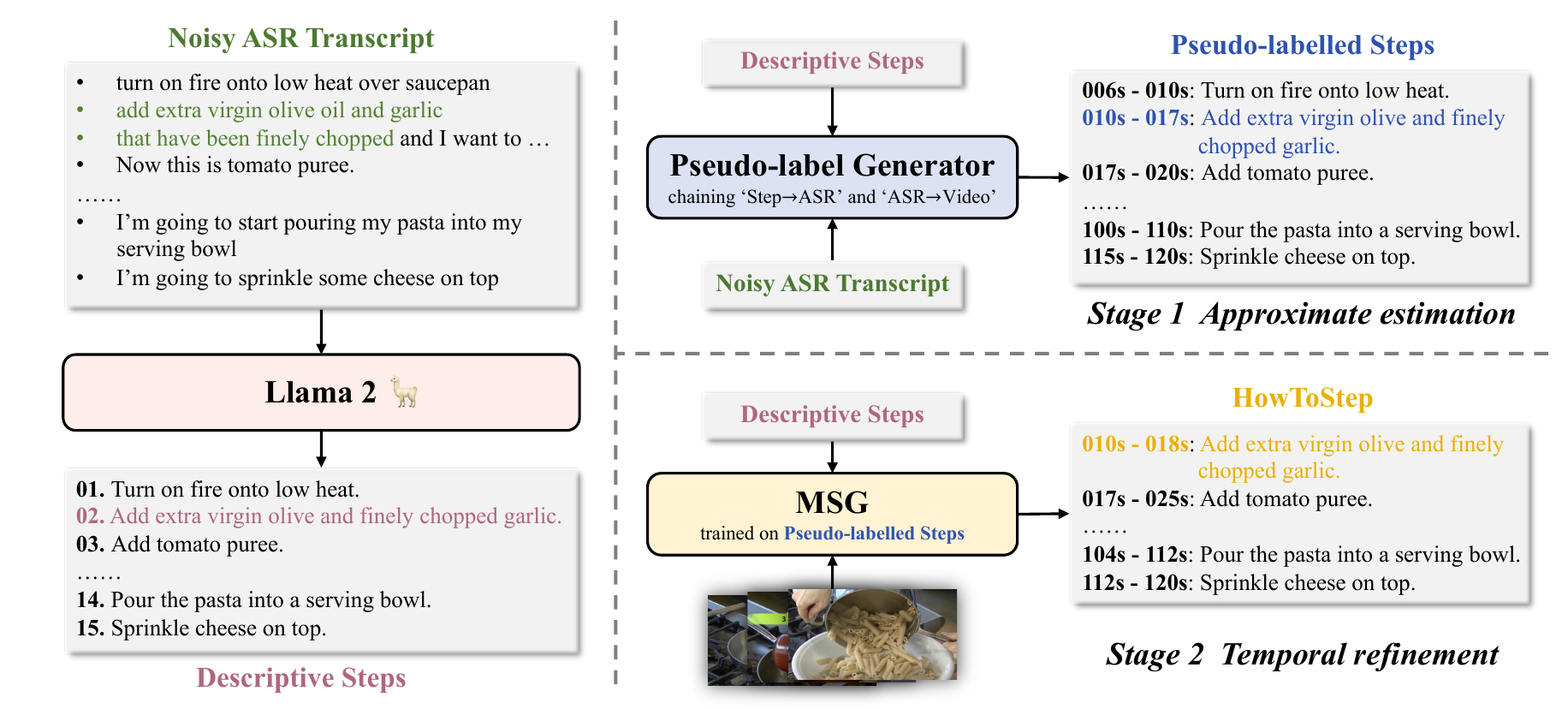}
   \caption{Schematic illustration of the proposed pipeline to summarizing noisy ASR transcripts into descriptive steps \textit{(left)}, while determining the start-end timestamp in the video \textit{(right)}. We utilize the Large Language Model (LLM) to summarize the narrations from ASR transcripts into descriptive steps. Afterwards, we roughly get the pseudo-label by chaining the `steps$\rightarrow$ASR' similarity and `ASR$\rightarrow$video' timestamp to train our multi-sentence grounding network \textbf{\modelname} in \textbf{\textit{Stage 1}}.
   Lastly, we use the trained model to refine the timestamp of the generated steps in \textbf{\textit{Stage 2}},
   resulting in an extra training source for multi-sentence grounding, named \textbf{HowToStep}.
   }
   \label{fig:pipeline}
\end{figure*}

\myparagraph{Prompting LLM to Summarize Procedural Steps}
We start by splitting the complete transcript for each video into $M$ segments, each segment $G_{i}$ consists of around 10 sentences. Next, we prompt a Llama2-7B~\cite{touvron2023llama} to summarize the transcript into concise steps that describe the main actions presented in the video, as well as filtering the colloquial sentences in the speech. We refer the readers to the complete prompt in \textbf{Appendix}. 

Formally, let $\Theta, \kappa$ refer to LLM and the prompt texts, and the steps are generated separately due to the limits of context length of LLM:
\begin{equation*}
    \mathcal{S} = \{\hat{G}_1, \dots, \hat{G}_M\},  \text{\hspace{4pt}}  
    \hat{G}_i = \Theta  (G_i;\kappa),  \text{\hspace{3pt}} \forall i \in [1, M]
    \label{eq:prompt}
\end{equation*}
where $M$ is the number of separated ASR transcript segments. 
$\hat{G}_i$ refers to the summarized procedure steps for each segment $G_i$. 
$\mathcal{S}$ denotes the complete sequence of generated steps for one instructional video.
As a result, we have transformed the ASR transcripts of around 370K videos~(a subset of HowTo100M selected by~\cite{han2022temporal}) into approximately 7M descriptive procedural steps. For comparison, the Wikihow knowledge base only contains 100k procedural steps from 14K instructional articles, according to the statistics in~\cite{mavroudi2023learning}. 

\myparagraph{Estimating Timestamp For Procedural Steps} 
Till this point, we propose to equip the summarised procedural steps with corresponding video timestamps, by mapping them back to the original narrations with sentence similarity.
Specifically, we compute the similarity score between $S$ generated steps and $N$ narrations of the transcript from each video, getting `steps$\rightarrow$transcript' matrix:
\begin{equation*}
    \mathbb{T}_{\mathcal{S}\mathcal{N}} =  \text{softmax}(g(\mathcal{S}) \cdot g(\mathcal{N})^T/\nu, \text{dim}=1) 
\end{equation*}
where $g(\cdot)$ is a pre-trained text encoder, $\nu$ is temperature, and $\mathbb{T}_{\mathcal{S}\mathcal{N}}\in \mathbb{R}^{S \times N}$ is the textual similarity matrix. The generated steps grounding score matrix can be computed by chaining the `steps$\rightarrow$transcript' matrix $\mathbb{T}_{\mathcal{S}\mathcal{N}}$
and `transcript$\rightarrow$video' matrix $\mathbb{Y}_{\mathcal{N}\mathcal{V}}$ (\ie, ASR timestamps).
\begin{equation*}
    \mathbb{Y}_{\mathcal{S}\mathcal{V}} =  \mathbb{T}_{\mathcal{S}\mathcal{N}} \cdot \mathbb{Y}_{\mathcal{N}\mathcal{V}}, \quad \mathbb{Y}_{\mathcal{S}\mathcal{V}} \in \mathbb{R}^{S \times T}
\end{equation*}
We set the timestamp with maximal score as the centre time $c_k$ for each step $k$, 
\begin{equation*}
    c_k = \mathop{\text{argmax}}_t \mathbb{Y}_{\mathcal{S}\mathcal{V}}^{[k, t]}
\end{equation*}
and then find start-end timestamp ($s_k, e_k$) from $c_k$ until the alignment score ($\mathbb{Y}_{\mathcal{S}\mathcal{V}}^{[k, s_k]}, \mathbb{Y}_{\mathcal{S}\mathcal{V}}^{[k, e_k]}$) are lower than the percentage $\zeta$ of the score of the center time~(\ie, $\mathbb{Y}_{\mathcal{S}\mathcal{V}}^{[k, c_k]} \times \zeta$) following~\cite{mavroudi2023learning}. 
The step whose maximal alignment score $\mathbb{Y}_{\mathcal{S}\mathcal{V}}^{[k, c_k]}$ is lower than a threshold $\epsilon_1$ will be regarded as unalignable and discarded.

\subsection{Multi-Sentence Grounding Model}
\label{subsec:network}

At this point, the start-end timestamps of the generated steps are obtained directly from weakly-aligned transcripts, which are naturally inaccurate, here, we propose a multi-sentence grounding model, termed as \textbf{Na}rrations / \textbf{S}teps to \textbf{V}ideo \textbf{A}ligner (\textbf{\modelname}), and a self-training strategy to refine these timestamps. Specifically, given an untrimmed long-form instructional video~(6.7 minutes on average), $\mathcal{X}=\{\mathcal{V}, \mathcal{J}\}$, where $\mathcal{V} = \{V_1, V_2, \dots, V_T\}$ refers to the frames of the video, and $\mathcal{J} = \{J_1, J_2, \dots, J_K\}$ denotes the $K$ textual sentences associated with the video. 
Our goal is to train a temporal multi-sentence grounding network that takes one video and multiple texts as inputs, and outputs a textual-visual alignment score matrix ($\hat{\mathbb{A}}\in \mathbb{R}^{K \times T}$) with the binary visual alignability ($\hat{y}\in \mathbb{R}^{K\times 2}$) for each sentence. Formally, it can be denoted as:
\begin{align}
 \{\hat{y}, \hat{\mathbb{A}} \} = \Psi_{\text{align}}(\Phi_{\text{v-enc}}(\mathcal{V}), \Phi_{\text{t-enc}}(\mathcal{J}))
\end{align}
In training, the ground truth label $\mathbb{Y}_{\mathcal{J}\mathcal{V}} \in \{0, 1\}^{K \times T}$ takes value $\mathbb{Y}_{\mathcal{J}\mathcal{V}}^{[k,t]} = 1$ only if $k$-th text depicts the scene of timestamp $t$ in the video, and zero otherwise. 

\begin{figure}[!t]
  \centering
   \includegraphics[width=0.95\linewidth]{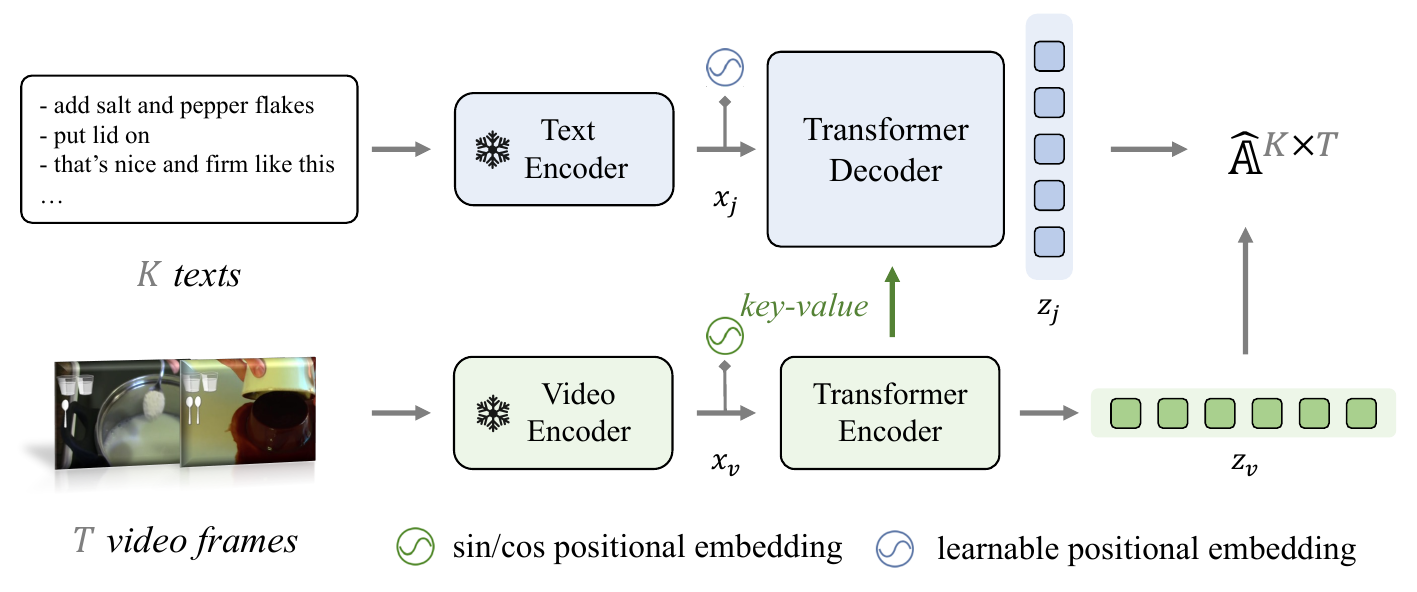}
   \caption{Schematic visualization of the proposed multi-sentence grounding network termed \textbf{\modelname}.
   The visual features are treated as key-value pairs while textual features as queries, to predict the alignment score matrix $\hat{\mathbb{A}}$ between video and texts.
   }
   \label{fig:arch}
\end{figure}

\noindent \textbf{Overall Architecture.} 
As shown in \cref{fig:arch}, we adopt a simple Transformer-based architecture, 
where visual features of each 1s clip are individually encoded and treated as key-value pairs, 
the textual features of narrations or steps are treated as queries.
The queries iteratively attend visual features with cross attention, 
and the alignment scores can be computed by textual-visual similarity, 
which emits high scores for any alignable sentences at their corresponding video timestamp. 
The following sections describe the full details.

\myparagraph{Visual-Textual Features} 
Given an instructional video 
with $K$ associated texts, 
we extract the visual and textual features with pre-trained backbones, 
\begin{equation*}
    x_v = f(\mathcal{V}) \in \mathbb{R}^{T\times C}, \quad  x_j = g(\mathcal{J})\in \mathbb{R}^{K\times C}
    \label{eq:backbone}
\end{equation*}
where $x_v$ refers to features of a video sequence, and $x_j$ denotes the textual features associated with the video. $C$ is the dimension of the feature vector. Note that, the resulting feature dimensions depend on the pre-trained backbone. 

We consider three popular pre-trained visual-language models, 
namely, the S3D-G backbone trained with MIL-NCE~\cite{xie2018rethinking, miech2020end}, CLIP-ViT/L-14 trained with InfoNCE~\cite{oord2018representation, radford2021learning, dosovitskiy2020image}, and InternVideo-MM-L14 trained with masked video
reconstruction and multi-modal contrastive loss~\cite{wang2022internvideo, wang2023videomae}. The ablation studies about these pre-trained backbones are presented in Sec. \ref{subsec:exp_ablation}.





\myparagraph{Multi-Sentence Grounding Module} As shown in \cref{fig:arch}, after extracting the visual and textual features independently, we project both features into the same dimension, fuse multimodal information with Transformer, and then predict the alignment score matrix between video and texts.


\myparagraph{(1) Feature Projection} 
After computing visual-textual features with pre-trained frozen models, we adopt one linear layer to project the features into the embedding with the same dimension $D$. In terms of positional encoding, we add sin/cos positional encoding to the visual features:
\begin{equation*}
    h_v = \phi_v(x_v)+p_v, \quad h_j = \phi_j(x_j)+\mathbb{I}_{\mathcal{J}=\mathcal{N}}\cdot p_j
    \label{eq:pe}
\end{equation*}
where $h_v \in \mathbb{R}^{T \times D}, h_j \in \mathbb{R}^{K \times D}$. $\phi_v, \phi_j$ refer to different projection heads for the features of video and sentences. $\mathbb{I}_{\mathcal{J}=\mathcal{N}}$ is the indicator function which takes value 1 only when input texts $\mathcal{J}$ are ordered narrations $\mathcal{N}$, otherwise zero. $p_v, p_j$ denote positional encoding for visual and textual features respectively. 

\myparagraph{(2) Visual-Textual Feature Fusion} 
The visual features are processed with a temporal aggregator,
followed by a grounding module, expressed as:
\begin{equation*}
    o_v = \Phi_{\text{temp-agg}}(h_v), \quad o_j = \Psi_{\text{temp-ground}}(o_v, h_j)
    \label{eq:transformer}
\end{equation*}
where $o_v \in \mathbb{R}^{T \times D}, o_j \in \mathbb{R}^{K \times D}$, 
$\Phi_{\text{temp-agg}}(\cdot)$ refers to a temporal aggregator with three Transformer Encoder layers. $\Phi_{\text{temp-ground}}(\cdot)$ denotes a temporal grounding module, consisting of three Transformer Decoder layers, where visual features act as key-value pairs and textual features act as queries.

\myparagraph{(3) Alignment Prediction}
To get the alignment score matrix between video and texts, 
we project the encoder and decoder outputs into the same dimension,
\begin{equation*}
    z_v = \varphi_v(o_v) \in \mathbb{R}^{T \times d}, \quad z_j = \varphi_j(o_j) \in \mathbb{R}^{K \times d}
    \label{eq:proj}
\end{equation*}
and then compute the alignment score matrix:
\begin{align*}
    \hat{\mathbb{A}}^{[k,t]} = \frac{{z_j^k} \cdot {z_v^t}^T}{\left\lVert z_j \right\rVert \cdot \left\lVert z_v \right\rVert} \in [0, 1]
    \label{eq:similarity}
\end{align*}
where $\hat{\mathbb{A}} \in \mathbb{R}^{K \times T}$ is the predicted alignment matrix. The higher value of $\hat{\mathbb{A}}^{[k,t]}$ means the $k$-th sentence is more likely to align with the scene of timestamp $t$.

\subsection{Self-training to Refine Generated Steps Grounding}
\label{subsec:self-training}

In this section, we describe the procedure to refine the start-end timestamp of generated procedural steps, by first training the proposed multi-sentence grounding model, and use it to update the labels.

\noindent \textbf{Training with Estimated Timestamps of Procedural Steps.}
Given the approximately estimated timestamps for the generated steps in \cref{subsec:llm}, denoted as 
$\mathbb{Y}_{\mathcal{J}\mathcal{V}}\in \{0, 1\}^{K \times T}$, we train the multi-sentence grounding network with a variant of the InfoNCE loss, 
following~\cite{han2022temporal, mavroudi2023learning}:
\begin{equation*}
    \mathcal{L}= -\frac{1}{K}\sum^K_{k=1}\log \frac{\sum_t \mathbb{Y}_{\mathcal{J}\mathcal{V}}^{[k,t]}\exp(\hat{\mathbb{A}}^{[k,t]}/\tau)}{\sum_t\exp(\hat{\mathbb{A}}^{[k, t]}/\tau)}
    \label{eq:infonce}
\end{equation*}
where $\hat{\mathbb{A}} \in \mathbb{R}^{K \times T}$ is the model's output alignment matrix, as explained in Sec.~\ref{subsec:network},
$\tau$ is a temperature hyper-parameter, and $k, t$ refer to $k$-th sentence and $t$-th timestamp in the video respectively. 

\noindent \textbf{Updating the Timestamps of Procedural Steps.}
We use the trained model to do inference on the whole set of procedural steps generated by LLM.
Specifically, when feeding video and a set of descriptive steps as model input, 
the proposed \textbf{\modelname} model outputs the alignment matrix~($\hat{\mathbb{A}}$). We take the timestamp of the maximal alignment score as the start time for each step, and the duration is a constant $\vartriangle_{\text{sec}}$, following~\cite{chen2021multimodal, shvetsova2023howtocaption}. Similarly, the generated step with the maximal alignment score lower than a threshold $\epsilon_2$ will be discarded. Finally, we obtain a dataset consisting of aligned descriptive steps, named \textbf{HowToStep}. In practice, we observe that although our \textbf{\modelname} model is trained on noisy pseudo-labels~(\ie, relying on weakly-aligned transcripts, and getting steps-transcript similarity from imperfect pre-trained text encoder), the model tends to learn the alignment patterns from procedural step to videos before overfitting to the noises, as also being observed in~\cite{zhang2021understanding}. Note this refined process (i.e., self-training) can be repeated multiple rounds. To balance computation cost and performance improvement, we choose only one round as default.
More details for generating the dataset are presented in ablation studies in \cref{subsec:exp_ablation} and \textbf{Appendix} as well.

\begin{figure*}[!t]
  \centering
   \includegraphics[width=\linewidth]{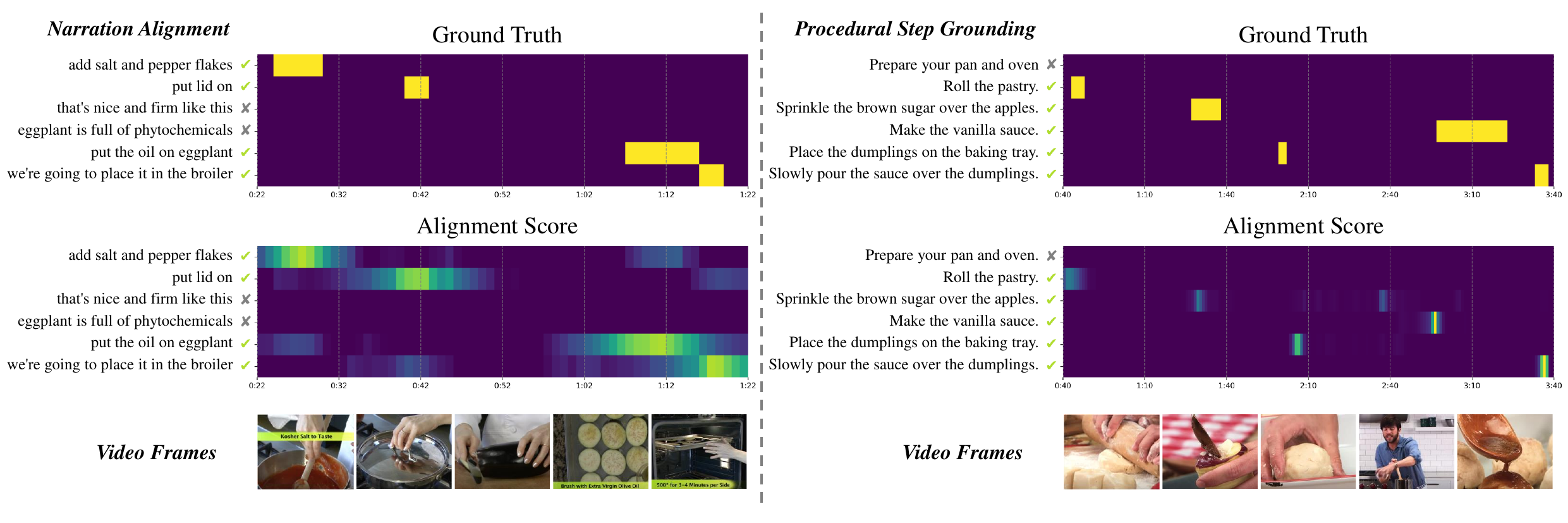}
   \caption{Qualitative examples of manually annotated visually-aligned text-to-video alignment matrix $\mathbb{Y} \in \{0,1\}^{K \times T}$ and the learned text-to-video alignment score matrix $\hat{\mathbb{A}} \in \mathbb{R}^{K \times T}$ of the model output for samples from HTM-Align~\textit{(left)} and HT-Step~\textit{(right)}. The ground truth timestamps of the example on HT-Step are labelled manually. Note that the temporal density and order of texts are quite different between the two tasks.
   }
   \label{fig:qualitative}
\end{figure*}

\myparagraph{Relation to Existing Work}
We note that recent works \cite{chen2022weakly, mavroudi2023learning, afouras2024ht} aim to ground steps collected from the only task-related articles in the external knowledge base~(\ie, Wikihow) within videos from HowTo100M as shown in \cref{fig:example} \textit{(middle)}, while we try to ground the procedural steps summarized from the ASR transcript of the \emph{same} video. This motivation has two natural advantages: (i) the procedural steps generated per task are more diverse than those from Wikihow (\eg, 74 different videos are showing how to make pumpkin puree in HowTo100M, while only one article of the same task is collected in Wikihow); 
(ii) the generated steps within the video are more likely to be alignable to the video itself than those externally sourced Wikihow steps intrinsically.

\subsection{Downstream Applications}
\label{subsec:app}


To demonstrate the effectiveness of the automatic pipeline for curating high-quality video-text datasets at scale, we consider using \textbf{HowToStep} for training and evaluation on multi-sentence grounding tasks. As shown in \cref{fig:qualitative},
we consider two variants, depending on the type of given texts $\mathcal{J}\in\{\mathcal{N}, \mathcal{S}\}$ (narrations or procedural steps), and differentiate the two targeted tasks as follows: 

\myparagraph{Narration Alignment~\cite{han2022temporal}}
$\mathcal{N}$arrations ($\mathcal{N}$) are textual sentences transcribed from the instructor's speech via the ASR system (around 100 sentences per video in HTM-Align~\cite{han2022temporal}), which means these sentences maintain a time order that can be utilized for visual grounding. Note that, some narrations are not visually alignable or not well-aligned with visual signals temporally as shown in \cref{fig:example} \textit{(left)}. However, the ASR transcripts with timestamps can be used as weakly-aligned labels $\mathbb{Y}_{\mathcal{N}\mathcal{V}}$ for training multi-sentence grounding network as previous work does~\cite{han2022temporal}.

\myparagraph{Procedural Step Grounding~\cite{mavroudi2023learning, zhukov2019cross}}
Procedural $\mathcal{S}$teps ($\mathcal{S}$) are collected from 
Wikihow~\cite{koupaee2018wikihow}, an external knowledge base with instructional articles explaining the procedures for completing certain tasks (around 10 steps per task in HT-Step~\cite{mavroudi2023learning}) or pre-defined action taxonomy (around 5 steps per video in CrossTask~\cite{zhukov2019cross}). Compared with narrations, the procedural steps do not necessarily have consistent temporal order as that happens in the video, and some steps may not even be present in the video at all. 

\section{Experiments}
\label{sec:experiments}

Here, we start by describing the datasets for training and evaluation, then present the implementation details and results for the multi-sentence grounding tasks.

\subsection{Datasets and Metrics}
Following the previous work~\cite{han2022temporal, mavroudi2023learning}, 
we perform dataset curation on a subset of the HowTo100M dataset,
and train our \textbf{\modelname} model with narrations from transcripts and steps from \textbf{HowToStep}. As for evaluation, we conduct procedural step grounding on HT-Step~\cite{mavroudi2023learning}, narration alignment on HTM-Align~\cite{han2022temporal}, 
and zero-shot action step localization on CrossTask~\cite{zhukov2019cross}.

\myparagraph{HTM-370K (Training)} 
The HowTo100M dataset~\cite{miech2019howto100m} is a large-scale instructional dataset crawled from YouTube, consisting of approximately 1.2M videos with ASR transcripts. Following previous work~\cite{han2022temporal,mavroudi2023learning}, 
we use HTM-370K for training, it contains videos from the Food \& Entertaining categories, consisting of 32\% of the videos of the entire HowTo100M dataset. 


\myparagraph{HowToStep (Training)} 
As introduced in~\cref{sec:method}, we construct this dataset for training, 
by transforming the original transcripts of HTM-370K into around $4$M ordered instructional steps with post-determined start/end timestamps for almost 340K videos after filtering.

\myparagraph{HTM-Align (Evaluation)} 
This benchmark is proposed by~\cite{han2022temporal} for evaluating narration alignment. It contains 80 videos from the HTM-370K as a holdout testing set. The authors have manually labelled the alignability for each narration and further aligned them to the visual signal with start/end timestamps if alignable. 
The metric on this dataset is Recall@1 (R@1), 
which means if the maximally matched timestamp for each alignable sentence model predicted falls into the ground truth temporal window, it is regarded as being successfully recalled. 
The recall score is computed as the ratio of successfully recalled sentences to all the alignable sentences.

\myparagraph{HT-Step (Evaluation)}
This benchmark~\cite{mavroudi2023learning} aims to evaluate the procedural step grounding. It contains manual annotations for 600 videos, specifically, for each video, the authors first collect activity steps from the related Wikihow article using the task name, \eg, Make Pumpkin Puree, and then annotate the temporal segment for steps alignable with the video. The metric on this dataset is the same as HTM-Align, namely, R@1.

\myparagraph{CrossTask (Evaluation)} 
In addition to benchmarks based on HowTo100M, 
we also adopt this established instructional video benchmark for zero-shot step localization. The CrossTask Dataset~\cite{zhukov2019cross} contains 4800 videos, which can be divided into 18 primary tasks and 65 related tasks. 
The videos in the primary tasks are annotated as steps with temporal segments from a predefined taxonomy. The metric is Average Recall@1 (Avg. R@1), which measures the recall over steps in videos for each task and averages the results. Following previous work~\cite{mavroudi2023learning}, we evaluate on a random set of 1850 videos from the primary tasks.

\begin{table}[t]
\footnotesize
  \centering
  \caption{\textbf{Comparison with the state-of-the-art }for narration alignment on HTM-Align, step grounding on HT-Step, and zero-shot action step localization on CrossTask. The results of TAN* are reproduced by \cite{mavroudi2023learning}. `\textbf{ZS}' refers to zero-shot.}
  \begin{tabular}{
  >{\arraybackslash}m{2.0cm}
  >{\centering\arraybackslash}m{1.8cm}
  >{\centering\arraybackslash}m{2.5cm}
  >{\centering\arraybackslash}m{2.5cm}
  }
    \toprule
    \multirow{2}{*}{\textbf{Method}}  
    &   \textbf{HT-Step}  & \textbf{HTM-Align}  & \textbf{CrossTask} (\textbf{ZS}) \\
    &  $\uparrow$ R@1 & $\uparrow$ R@1 & $\uparrow$ Avg. R@1 \\
    \midrule
    Zhukov~\cite{zhukov2019cross} & - & - & 40.5 \\ 
    UniVL~\cite{luo2020univl} & - & - & 42.0 \\
    TAN*~\cite{han2022temporal}  & 31.2  & 47.1 & - \\
    VINA~\cite{mavroudi2023learning} & 37.4 & 66.5 & 44.8 \\ 
    \midrule
    \textbf{Ours} & \textbf{46.4} & \textbf{71.6} & \textbf{46.7} \\ 

    \bottomrule
  \end{tabular}
  \label{tab:comparison}
\end{table}

\subsection{Implementation Details}

Overall, we investigate the effectiveness of three popular pre-trained visual-language models for constructing the whole pipeline, namely, the S3D-G, CLIP-ViT/L-14, and InternVideo-MM-L14, as described in \cref{subsec:network}. While exploring other factors, for example, the effect of ASR transcripts,
and the effect of incorporating descriptive steps during training, 
we use InternVideo-MM-L/14 by default, unless specified otherwise.
At training time, the temperature $\tau$ in our loss is 0.07. 
We use the AdamW~\cite{loshchilov2017decoupled} optimizer 
and train the model with an initial learning rate $10^{-4}$ and cosine decay for 12 epochs. 
When determining the start/end time for generated steps, 
the hyper-parameters for the 2-stage are $\zeta=0.7, \epsilon_1=0.20$, $\epsilon_2=0.8$, which are also discussed in the ablation study. 
We train one unified model for both narration alignment and procedural step grounding tasks, by setting the texts of one training batch to be either ordered, dense narrations or shuffled, sparse procedural steps. Complete implementation details are included in the \textbf{Appendix}.

\subsection{Main Results}


\myparagraph{Comparison with State-of-the-art}
We compare our best model with existing state-of-the-art approaches on three public benchmarks for multi-sentence grounding tasks. 
As shown in \cref{tab:comparison}, on the challenging HT-Step task, 
that aims to ground unordered procedural steps in videos, 
our model achieves 46.4\% R@1, leading to an absolute improvement of 9.0\%, over the existing state-of-the-art (37.4\%) achieved by VINA~\cite{mavroudi2023learning}.
On HTM-Align~\cite{han2022temporal}, which aligns narrations in the video, 
our method exceeds the SOTA model by 5.1\%. 
On CrossTask~\cite{zhukov2019cross}, where we need to align video frames and task-specific steps without finetuning, our method outperforms existing the state-of-the-art approach by 1.9\%,
demonstrating the effectiveness of the proposed pipeline for downstream tasks.

\subsection{Ablation Study}
\label{subsec:exp_ablation}
We explore the effects of multiple design choices in the proposed pipeline and evaluate them on both narration alignment and procedural step grounding tasks.

\myparagraph{Effect of Upgrading ASR System} 
To start with, we compare the original transcripts scrawled from YouTube, with that from the recent WhisperX~\cite{bain2022whisperx, radford2023robust}, as the weakly-aligned labels for training. Qualitatively, we do observe that WhisperX generates fewer punctuation errors, and gives higher accuracy of temporal boundaries in the ASR transcripts. As shown in \cref{tab:ablation_asr}, upgrading the ASR system indeed leads to noticeable performance improvement in narration alignment. However, in procedural step grounding, showing a marginal performance decrease, we conjecture this is because the gap of text style between the train set~(dense speeches) and test set~(sparse procedural steps) dominates the performance. 
In later sections, we use the WhisperX transcripts by default.

\myparagraph{Effect of Upgrading Visual-Textual Backbone} 
Here, we explore different visual backbones with the corresponding text encoder. As shown in \cref{tab:ablation_asr}, S3D-G exceeds CLIP ViT/L-14 in narration alignment but is inferior to the latter in step grounding. In general, InternVideo ViT/L-14 shows significant advantages on both tasks, attributed to the large pre-trained dataset and effective supervision in video representation learning, which is our default choice in the pipeline.

\begin{table}[!t]
\footnotesize
  \centering
  \caption{\textbf{Ablation of transforming ASR transcripts into descriptive steps with post-determined timestamps as the extra training source.}
  `W' denotes transcripts from WhisperX, and `S' denotes our proposed HowToStep dataset.} 
  \resizebox{0.78\textwidth}{!}{
  \begin{tabular}{
  >{\arraybackslash}m{3.8cm}
  >{\arraybackslash}m{2.0cm}
  >{\centering\arraybackslash}m{2.2cm}
  >{\centering\arraybackslash}m{2cm}
  }
    \toprule
    \multirow{2}{*}{Backbone} & \multirow{2}{*}{Training Text}  & \textbf{HT-Step} & \textbf{HTM-Align} \\
    &  & $\uparrow$ R@1 & $\uparrow$ R@1 \\
    \midrule
    CLIP-ViT/L-14~\cite{radford2021learning} 	 & W   &  32.2 & 58.7 \\
    CLIP-ViT/L-14	 & W + S  & 42.4 &  64.9  \\
    \midrule
    MIL-NCE S3D-G~\cite{miech2020end} & W   & 29.1 & 59.5 \\
    MIL-NCE S3D-G & W + S   & 40.9 &  63.0  \\
    \midrule
    InternVideo-ViT/L-14~\cite{wang2022internvideo} & W  & 34.7 & 70.0  \\
    \cellcolor{gray!15}InternVideo-ViT/L-14 & \cellcolor{gray!15}W + S  & \textbf{46.4} & \textbf{71.6}  \\
    \bottomrule
  \end{tabular}}
  \label{tab:ablation_transform}
\end{table}

\begin{table}[!t]
\footnotesize
  \centering
 \caption{\textbf{Ablation of choices in constructing the pipeline.} `Step-Video' means to determine the start/end time directly by computing visual-textual similarity, while `Step-ASR' means indirectly by chaining `steps$\rightarrow$transcript' similarity and `transcript$\rightarrow$video' timestamp. The texts used for training here are only the generated steps (`S').
  }
 \resizebox{0.65\textwidth}{!}{
  \begin{tabular}{
  >{\centering\arraybackslash}m{2.5cm}
  >{\centering\arraybackslash}m{0.9cm}
  >{\centering\arraybackslash}m{2cm}
  >{\centering\arraybackslash}m{0.9cm}
  >{\centering\arraybackslash}m{2.2cm}
  }
    \toprule
    \multicolumn{4}{c}{Method} & \multirow{2}{*}{\raisebox{4pt}{\textbf{HT-Step}}} \\[0.2em] \cline{1-4}
      Pesudo-label \rule{0pt}{\normalbaselineskip} & \rule{0pt}{\normalbaselineskip}$\epsilon_1$ & \rule{0pt}{\normalbaselineskip} Self-training & $\rule{0pt}{\normalbaselineskip} \epsilon_2$ &  $ \uparrow$ R@1 \\
    \midrule
        Step-Video & 0.20 & \xmark & - & 36.5 \\
        Step-Video & 0.20 & \cmark & 0.8 & 36.9 \\ \midrule
        Step-ASR  & 0.15 & \xmark & - & 35.3 \\  
        Step-ASR  & 0.20 & \xmark & - & 36.0 \\  \midrule 
        Step-ASR  & 0.15 & \cmark & 0.7  & 41.4 \\
        Step-ASR  & 0.15 & \cmark & 0.8  &  41.7  \\ \midrule
        Step-ASR  & 0.20 & \cmark & 0.7  & 42.3 \\
        \cellcolor{gray!20}Step-ASR  & \cellcolor{gray!20}0.20 & \cellcolor{gray!20}\cmark & \cellcolor{gray!20}0.8  & \textbf{43.7} \\
    \bottomrule
  \end{tabular}}
  \label{tab:ablation_timestamp}
\end{table}

\myparagraph{Effect of the Proposed Dataset} 
We validate the effectiveness of using this pipeline to transform noisy ASR transcripts into descriptive steps, with the post-determined temporal segments as an extra training source for multi-sentence grounding. As shown in \cref{tab:ablation_transform}, on all three backbones, the generated dataset is effective for both narration alignment and step grounding. Notably, the average improvement exceeds 10\% in HT-Step implies that the generated dataset are indeed more descriptive and well-aligned. For narration alignment, our generated steps add more diversity for training, thus leading to better performance.

\myparagraph{Ablation of Options in Constructing the Pipeline} 
We investigate the choices and hyper-parameters to generate the extra training dataset~(HowToStep) described in \cref{sec:method}. 
As shown in \cref{tab:ablation_timestamp}, 
the difference in determining the timestamp of the generated steps by directly computing the video-step similarity matrix, and indirectly by chaining `steps$\rightarrow$transcript' similarity and `transcript$\rightarrow$video' labels is not obvious for the first stage, while becomes significant when using the pseudo-label of the first stage for the self-training in the second stage, there is a large gap between the two methods. We find that using the video-step similarity matrix directly will make the model learn a trivial solution (\ie, identity mapping), while indirectly obtained pseudo-label can let the model learn the alignment patterns as analyzed in \cref{subsec:self-training}. In addition, we choose the best thresholds to generate the final dataset according to \cref{tab:ablation_timestamp}.

\begin{figure*}[!t]
\centering
\begin{minipage}{0.5\textwidth}
    \scriptsize
    \centering
    \captionof{table}{\textbf{Ablation study for visual-textual backbones and ASR systems.} Here we only use the weakly-aligned ASR transcripts to train our model.}
    \label{tab:ablation_asr}
    \begin{tabular}{
  >{\arraybackslash}m{1.7cm}
  >{\centering\arraybackslash}m{1.5cm}
  >{\centering\arraybackslash}m{1.2cm}
  >{\centering\arraybackslash}m{1.7cm}
  }
    \toprule
    \multirow{2}{*}{Backbone}  & \multirow{2}{*}{ASR System}   & \textbf{HT-Step} & \textbf{HTM-Align} \\
     & &  $\uparrow$ R@1 & $\uparrow$ R@1 \\
    \midrule
    CLIP &	WhisperX  &  32.2 & 58.7 \\
    S3D & WhisperX & 29.1 & 59.5 \\
    \cellcolor{gray!15}InternVideo & \cellcolor{gray!15}WhisperX  & 34.7 & \textbf{70.0}  \\
    InternVideo & Youtube  & \textbf{35.7} & 61.3  \\
    \bottomrule
  \end{tabular}
\end{minipage}
\hspace{0.05\textwidth}
\begin{minipage}{0.4\textwidth}
    \scriptsize
    \centering
    \captionof{table}{\textbf{Manual check of the dataset quality.}  HTM-370k and HowToStep are the datasets before and after the processing in our proposed pipeline, respectively.}
    \label{table:quality}
    \begin{tabular}{
    >{\arraybackslash}m{1.7cm}
  >{\centering\arraybackslash}m{1.1cm}
  >{\centering\arraybackslash}m{1.7cm}
  }
        \toprule
        \multirow{2}{*}{Dataset}  &   Alignable  & Well-aligned \\
        &  $\% \uparrow$ & $\% \uparrow$ \\
            
            \midrule
            HTM-370K & 30.1 & 21.9 \\
            HowToStep & \textbf{60.6} & \textbf{52.5} \\
            \bottomrule
        \end{tabular}
\end{minipage}
\end{figure*}

\subsection{Manual Check}
To evaluate the quality of our proposed dataset, we have randomly sampled 10 videos with a total of 853 sentences for a manual check, focusing on the proportion of steps in the dataset that are (i) visually alignable, (ii) and well-aligned with correct temporal boundaries. We compare the results before and after using our pipeline, as shown in \cref{table:quality}, demonstrating the significant quality improvement.

\section{Conclusion}
\label{sec:conclusion}
To conclude, we have established an automatic pipeline for constructing a high-quality video-text dataset for multi-sentence grounding in large-scale instructional videos. 
We have investigated the factors potentially affecting performance, including upgrading the ASR system, transforming the noisy ASR transcripts into descriptive steps by LLMs as an extra training source, and proposing a simple Transformer-based model to refine the temporal windows for each step.
When evaluating three public benchmarks of multi-sentence grounding tasks, our method surpasses the existing state-of-the-art methods by a significant margin.

\section*{Acknowledgements}
This work is supported by National Key R\&D Program of China (No.2022ZD0161400).

\bibliographystyle{splncs04}
\bibliography{main}

\clearpage

{
    \centering
    \large
    \textbf{Supplementary Materials} \\
}


\section{Additional Implementation Details}

\subsection{Architecture Details}
In this section, we provide more details on the visual-language backbones, and the grounding module, as introduced in Sec.~3.2~(main paper).

\myparagraph{Visual Backbone} 
We adopt S3D-G pre-trained with MIL-NCE, CLIP ViT/L-14 pre-trained with InfoNCE and InternVideo-MM-L-14 pre-trained with reconstruction and contrastive loss. 
For S3D-G, the procedure begins by decoding the original video at 16 frames per second (fps), followed by resizing it to ensure the shorter dimension is 256 pixels. After resizing, the frames are then center-cropped to a resolution of $224 \times 224$ before being inputted into the S3D-G. Each 16-frame video clip cropped by a non-overlapping temporal window is fed into the S3D-G architecture, resulting in one feature~(512-d) per second.
For CLIP and InternVideo, the original video is first center cropped to a shorter side and then resized to $224 \times 224$ resolution. 
When using CLIP, we decode the video into 1fps and extract one visual feature~(768-d) per second with OpenAI's ViT/L-14 model~\cite{dosovitskiy2020image}. 
As for InternVideo, We decode the video into 8fps and feed it into the InternVideo-MM-L14 model with an 8-frame non-overlapping temporal window, obtaining one visual feature~(768-d) per second.

\myparagraph{Textual Backbone}
The text encoder associated with S3D-G adopts a Bag-of-word~(BoW) model based on Word2Vec embeddings pre-trained on GoogleNews. 
Each sentence is tokenized, truncated under 32 tokens, and then encoded with the text encoder associated with the S3D-G video encoder~\cite{miech2020end}.
Specifically, the sentence is tokenized and then converted into word vectors~(300-d) through an embedding layer trained on Google News. Following this, word vectors from the same sentence will be projected into 512-d vectors, and turned into one sentence vector through maxpooling. 
For the text encoder of CLIP and InternVideo, it takes a maximum of 77 tokens for each sentence.
Each token is passed to an embedding layer and added with positional encoding. After being encoded by the Transformer blocks,  we take sentence features from the \texttt{<eot>} embedding for each sequence.

\myparagraph{Grounding Modules} For multi-sentence grounding modules in our proposed model, we use 3-layer Transformer encoder blocks and 3-layer decoder blocks with 8-head attention mechanisms.
The model dimension, denoted as $D$, is set to 256, while the projection dimension, represented as $d$, utilized for computing the cosine similarity, is 64. At training time, we will truncate the video, whose duration is longer than 1200 seconds, and set the batch size as 8. 
Our model is trained on a single GPU (NVIDIA GeForce RTX 3090) for approximately 20 hours on the whole training data (WhisperX + HowToStep).

\subsection{Dataset Details}

\begin{figure*}[!t]
  \centering
  \hspace{-10pt}
  \begin{subfigure}{0.4\textwidth}
    \includegraphics[width=\linewidth]{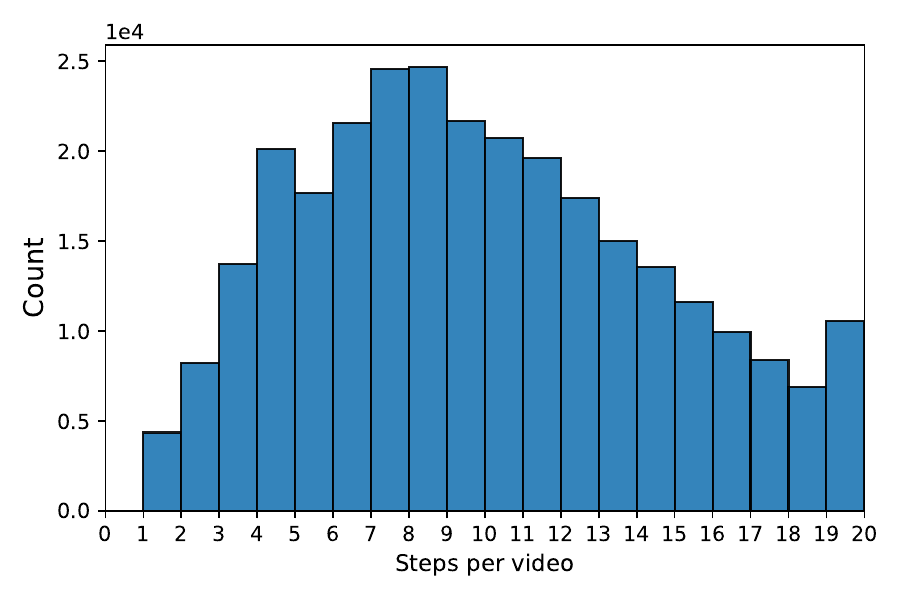}
    \caption{The distribution of number of steps per video.}
    \label{fig:steps}
  \end{subfigure}
  \hspace{12pt}
  \begin{subfigure}{0.4\textwidth}
    \includegraphics[width=\linewidth]{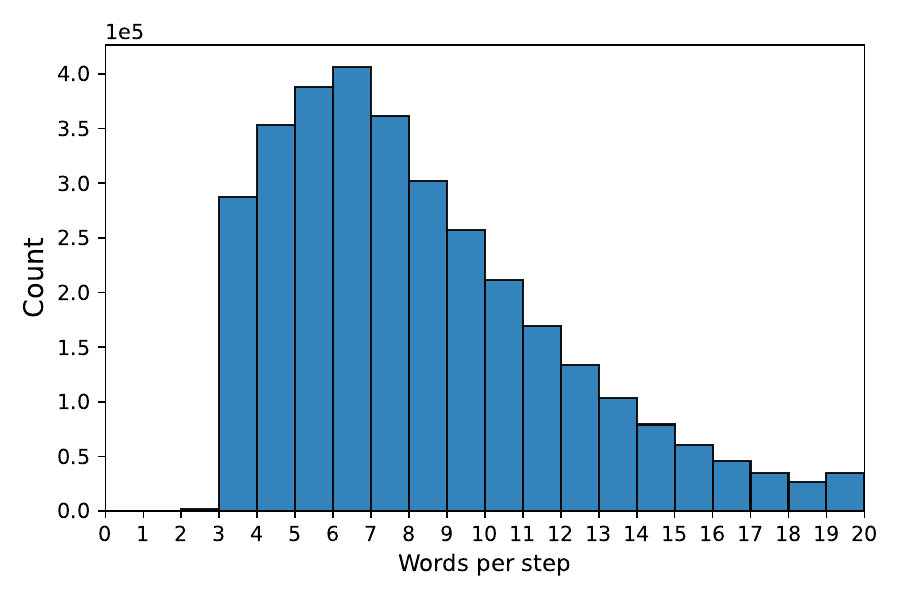}
    \caption{The distribution of number of words per step.}
    \label{fig:words}
  \end{subfigure}


  \begin{subfigure}{\textwidth}
    \centering
    \includegraphics[width=0.8\linewidth]{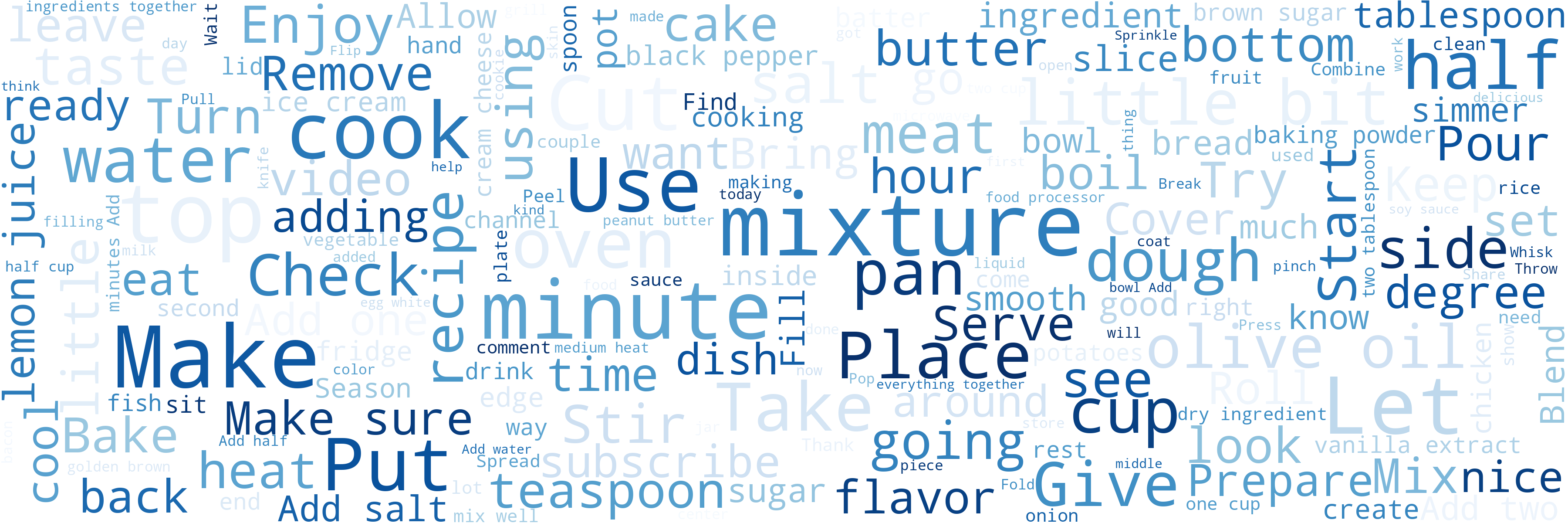}
    \caption{The keyword cloud of HowToStep.}
    \label{fig:keyword}
  \end{subfigure}

  \caption{Statistics visualization of HowToStep. We provide various statistics for a qualitative overview of the dataset.}
  \label{fig:statistics}
\end{figure*}

\begin{figure}[htbp]
\centering
\begin{subfigure}{.47\columnwidth}  
  \centering
  \includegraphics[width=\linewidth]{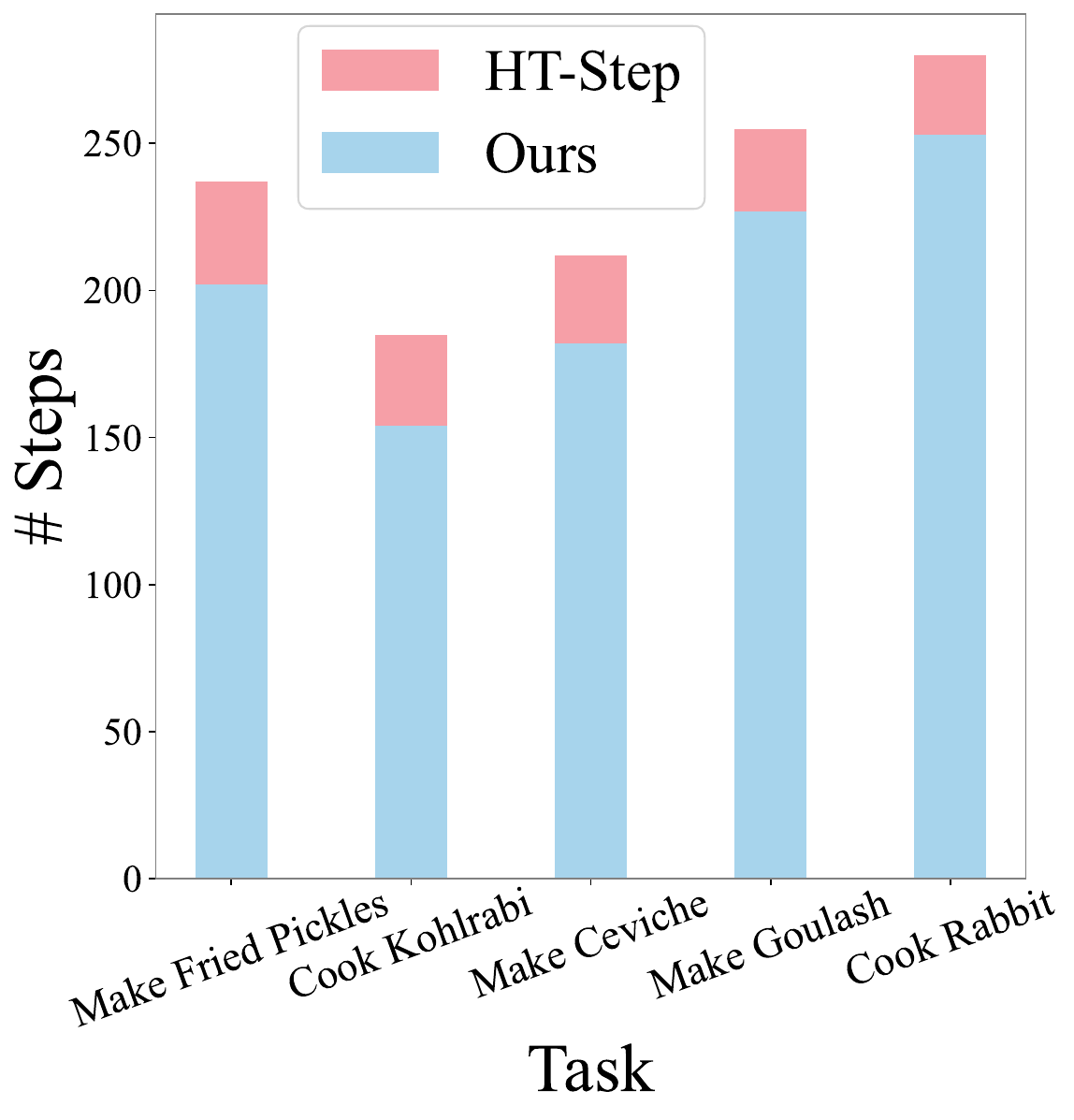}  
  \caption{Number of aligned steps across the tasks with most Wikihow steps.}
  \label{fig:sub1}
\end{subfigure}%
\hfill
\begin{subfigure}{.45\columnwidth}
  \centering
  \includegraphics[width=\linewidth]{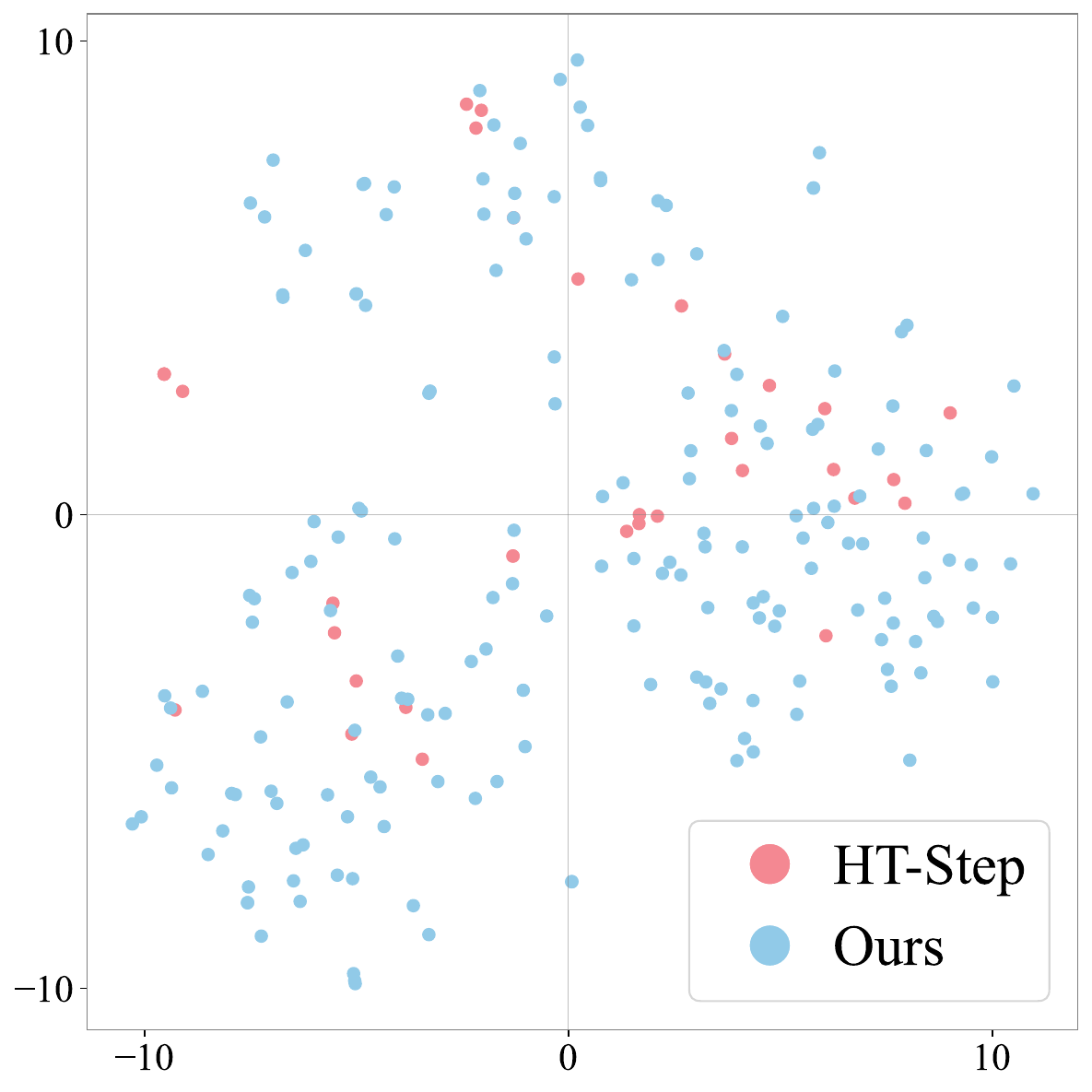}
  \caption{Visualization of sentence embeddings of aligned steps from the `\textit{How To Make Fried Pickles}' task.}
  \label{fig:sub2}
\end{subfigure}
\caption{Controlled comparison about the scalability and diversity of the procedural steps on our proposed dataset with HT-Step.}
\label{fig:diversity}
\end{figure}

\myparagraph{HowToStep (Training)}  
In our final HowToStep, the average number of steps ~(sentences) per video stands at 10.6, with an average word count per step amounting to 8.0. As shown in \cref{fig:steps} and \cref{fig:words}, we plot the distribution of steps per video and words per step. 
In addition, we present a word cloud to show the descriptions of the summarized steps in HowToStep~(\cref{fig:keyword}). 
To validate the diversity of procedural steps in our dataset, we conduct a quantitative comparison with the latest work [2] in \cref{fig:sub1},
by comparing the number of aligned steps within the same set of videos. On five tasks that have the most Wikihow steps, our step count is \textbf{11.2} times more than [2] on average. Additionally, using t-SNE to reduce sentence embedding dimensions for the 'How To Make Fried Pickles' task, our steps are semantically more diverse, as shown in \cref{fig:sub2}.  

\myparagraph{HTM-Align (Evaluation)} Given an instructional video from HowTo100M with its narrations with start-end timestamps from the YouTube ASR transcript, the annotator determines whether each narration is alignable with the video (\ie, visually related) and adjusts the ground truth temporal window to cover the visual content if the narration is alignable. We employ the metric, Recall@1, to evaluate whether the
predicted timestamp of the \textit{narrations are alignable} with the video falls into the ground truth temporal window~\cite{han2022temporal}.

\myparagraph{HT-Step (Evaluation)} Given the video from HowTo100M with the task name (\eg, Make Pumpkin Puree) and the recipe steps from the corresponding Wikihow article, the annotator decides whether the video is relevant to the task. If the video is relevant to the task of steps, the annotator will mark all the instances of the steps with a temporal window. The test set contains 600 videos, with 5 videos per task and the metric is Recall@1 as well. The complete statistics for this dataset can be found in \cite{mavroudi2023learning}.

\myparagraph{CrossTask (Evaluation)} The CrossTask dataset consists of 4800 videos from 18 primary activities and 65 related activities. The videos of primary activities are annotated with steps from a predefined taxonomy of 133 atomic steps (\eg, add onion, add taco) and the corresponding temporal windows in the video. For the step localization task on this dataset, the metric is Average Recall@1, which means computing Recall@1 for steps of videos from each primary activity and averaging the results of different activities. 
Following previous work~\cite{mavroudi2023learning}, we report the average results over 20 random sets of 1850 videos from 18 primary activities.

\subsection{Task Details}
For the narration alignment task, the texts are in the form of narrations in videos, with strong temporal correlations. The consecutive sentences in narrations typically follow a temporal order. 
However, in procedural step grounding task, the texts are in the form of steps collected from a knowledge base, namely, WikiHow. 
Due to the discrepancy between the knowledge base and the instructional videos, the order of steps in the knowledge base may differ from the order of actions in the videos. 
Therefore, the temporal order of consecutive sentences is not fixed.
Due to the different temporal attributes of the texts in the two tasks, as described in Sec.~3.2, we have delineated an indicator to ascertain the inclusion of learnable positional encodings for textual features. 
Specifically, we have defined a token that denotes the type of text. 
When the token is denoted as \textit{<narration>}, signifying that the input bears pronounced temporal correlations akin to a narration, we integrate positional encodings and, during training, input all the texts of one video into the grounding module in their temporal order. Conversely, when the token is represented as \textit{<step>}, indicating that the input text does not follow a temporal sequence, we refrain from incorporating positional encodings and input the texts of a single video into the grounding module in a shuffled sequence for training purposes.
Throughout the training phase, we randomly allocate the type-token of the training samples as \textit{<narration>} with a 50\% probability, and as \textit{<step>} with the remaining 50\% probability.  
During the inference stage of the narration alignment task, we classify the type-token for all samples as \textit{<narration>}; conversely, in the inference stage of the procedural step grounding task, we assign the type-token for all samples as \textit{<step>}. This methodology allows us to achieve both tasks with a unified set of weights.

\begin{table*}[!t]
  \centering
  \small
  \resizebox{0.9\textwidth}{!}{
  \begin{tabular}{m{9cm}c|c}
    \toprule
    \multirow{2}{*}{Prompt} & Two-Stage  & \textbf{HT-Step} \\
      & Determination &  $\uparrow$ R@1 \\
    \midrule
    I will give you an automatically recognized speech with timestamps from a video segment that is cut from a long video. Write a summary for this video segment. Write only short sentences. Describe only one action per sentence. Keep only actions that happen in the present time. \textbf{Begin each sentence with an estimated timestamp}. Here is this automatically recognized speech: $\langle$\textbf{\textit{timestamp + ASR transcript}}$\rangle$ & \xmark & 19.6 \\ \midrule
    I will give you an automatically recognized speech with timestamps from a video segment that is cut from a long video. The speaker in the video is teaching the audience to do something. Your task is to summarize the key steps in order. Each step should be short and concise phrase. Do not output colloquial sentences in the speech. \textbf{Output only the numbered key steps without timestamps}. Here is this automatically recognized speech: $\langle$\textbf{\textit{timestamp + ASR transcript}}$\rangle$ & \cmark & 41.8 \\ \midrule
    \cellcolor{gray!15} I will give you an automatically recognized speech from a video segment that is cut from a long video. The speaker in the video is teaching the audience to do something. Your task is to summarize the key steps in order. Each step should be short and concise phrase. Do not output colloquial sentences in the speech. Describe only one action per sentence. Output the numbered key steps. Here is this automatically recognized speech: $\langle$\textbf{\textit{ASR transcript}}$\rangle$ & \cellcolor{gray!15} \cmark & \cellcolor{gray!15} 43.7 \\
    \bottomrule
  \end{tabular}}
  \caption{\textbf{Ablation of prompts.} We experiment with various prompts to guide the LLM in generating descriptive steps on HTM-370K for training. We apply an identical temporal grounding approach, train the model only on the resulting subset, and then evaluate on HT-Step. 
  }
  \label{tab:ablation_prompt}
\end{table*}

\section{Extra Ablations}
\myparagraph{Ablation on dataset construction} 
In this section, we perform the ablations on HTM-370K, concerning the prompt for guiding the LLM in generating descriptive steps and the methodology for equipping the summarised procedural steps with corresponding video timestamps.
A straightforward method to acquire descriptive steps and their corresponding video segments is to feed both ASR transcripts and their respective timestamps into a LLM, prompting the LLM to simultaneously generate descriptive steps along with their associated timestamps. Specifically, we utilize the format $\langle$\textit{timestamp + ASR transcript}$\rangle$ as input, employing the prompt in the first row of \cref{tab:ablation_prompt} to guide the LLM in generating descriptive steps, along with timestamp for each step. However, given that timestamps from ASR transcripts may not be aligned, and LLM cannot rectify this misalignment based solely on text, the steps generated by this method cannot endow the model with proficient multi-sentence grounding capabilities. 
An alternative approach involves the LLM solely undertaking the task of step generation, followed by determining the timestamp for each step through the two-stage determination procedure proposed in Sec.~3.1 and Sec.~3.3. Employing this approach, we devised two prompts, as delineated in the second and third rows of \cref{tab:ablation_prompt}. The distinction lies in that for the second row, we provide the corresponding timestamp extracted from the ASR, with the intention of guiding the LLM to utilize temporal information for generating the steps; whereas, in the third row, we omit any input of timestamp information.
We use the same hyper-parameters in the 2-stage temporal grounding procedure for generated steps and then train on the same model to get the results on HT-Step. 
From \cref{tab:ablation_prompt}, it is evident that steps generated by the LLM without incorporating timestamps result in improved grounding capabilities for the model. This could be attributed to inherent misalignment within the timestamps extracted from ASR transcripts, which may inadvertently misguide the inference process of the LLM. On the other hand, when timestamps are included in the input, the outcomes generated by the LLM frequently contain timestamp information, despite restrictions imposed in the prompts against this.
Hence, we ultimately choose the prompt in the third row of \cref{tab:ablation_prompt} to generate descriptive steps.

\myparagraph{Ablation on LLM choice in proposed pipeline}
In this section, we performed an ablation study on roughly 10\% of the total training dataset, employing different LLMs, namely Llama-2, Mistral, and Llama-3. As shown in \cref{table:ablations_LLM}, Mistral and Llama-2 demonstrated similar performance, while Llama-3 exhibited the best performance. We adopt Llama-2, but the quality of data generated by our proposed pipeline can continuously improve with the advancement of LLMs.

\begin{table}[ht]
    \footnotesize
    \centering
        \resizebox{0.6\linewidth}{!}{
        \begin{tabular}{ccc}
        \toprule
            LLM & \textbf{HT-Step} R@1 $\uparrow$  & \textbf{HTM-Align} R@1 $\uparrow$ \\
            \midrule
            Llama-2 & 35.5 & 34.9 \\
        Mistral & 35.1 & 34.7 \\
        Llama-3 & \textbf{35.7} & \textbf{36.7} \\
            \bottomrule
        \end{tabular}}
        \caption{\textbf{Ablation of LLMs in the proposed pipeline.} Llama-2, Mistral, and Llama-3 correspond to Llama-2-7B-chat, Mistral-7B-Instruct-v0.2, and Llama-3-8B-Instruct, respectively.}
        \label{table:ablations_LLM}
        
\end{table}

\myparagraph{Ablation on temporal grounding} 
As mentioned in Sec.~3.1 and Sec.~3.3, we propose a two-stage method to determine the start/end timestamp for each generated descriptive step and we can repeat the refined process (i.e., self-training) multiple iterations. 
In \cref{tab:ablation_generate} we ablate the complete design choices for the first iteration of temporal refinement, including the filtering threshold, the position of maximal alignment score in each step segment, the duration of time allocated to each step.
Setting the maximum similarity as the starting timestamp of the step with a constant duration brings the best performance for the first interation refinement.
Subsequently, we conduct ablation study on the iterations of self-traininig, with each refinement round considering various filtering thresholds, positions of maximal alignment score, and step durations as what we do in Table 8. Consequently, we present the best result for each iteration in \cref{table:ablations_iterations}. It is evident that performing two iterations of temporal refinement yields the best performance. However, it is worth noting that in all the aforementioned experiments, we chose only one round as the default to balance computation cost and performance improvement.

\begin{table}[!t]
  \centering
  \small
  \begin{tabular}{@{}cc|ccc|ccc|ccc@{}}
    \toprule
    \multirow{3}{*}{Method} & $\epsilon_2$ & 0.6 & 0.7 & 0.8 & 0.8 & 0.8 & 0.8 & 0.8 & 0.8 & 0.8 \\ 
     & pos & center & center & center & start & start & start & center & center & center \\
     & $\Delta_{sec}$ & \xmark & \xmark & \xmark & 6 & 8 & 10 & 6 & 8 & 10 \\  \cmidrule(r){1-11} 
    \multicolumn{2}{c|}{\makecell[c]{\textbf{HT-Step} \\ $\uparrow$ R@1}} & 37.6 & 37.5 & 38.2 & 43.1 & \textbf{43.7} & 43.0 & 38.0 & 37.2 & 37.8 \\
    \bottomrule
  \end{tabular}
  \caption{\textbf{Ablation of choices in first iteration temporal refinement.} We use Step-ASR relation to obtain pseudo-labels. The texts used for training the model here are only HowToStep (S). \xmark~means we do not fix the duration for each generated step, but find the start-end timestamp from the center point as we do in the approximate estimation stage.
  }
  \label{tab:ablation_generate}
\end{table}

\begin{table}[ht]
    \footnotesize
    \centering
        \resizebox{0.8\linewidth}{!}{
        \begin{tabular}{ccccccc}
        \toprule
            \multirow{2}{*}{Iters} & \multirow{2}{*}{thresh} & \multirow{2}{*}{pos} & \multirow{2}{*}{$\Delta_{sec}$} & Training  & \textbf{HT-Step} & \textbf{HTM-Align} \\
        & & & & Text & $\uparrow$ R@1 & $\uparrow$ R@1 \\
            \midrule
            \xmark & 0.2 & center & \xmark & S & 36.0 & 44.0 \\
            \midrule
            1 & 0.8 & start & 8 & S & 43.7 & 46.7 \\
            1 & 0.8 & start & 8 & W + S & 46.4 & 71.6 \\
            \midrule
            2 & 0.7 & center & \xmark & S & 44.8 & 52.9 \\
            \cellcolor{gray!15}2 & \cellcolor{gray!15}0.7 & \cellcolor{gray!15}center & \cellcolor{gray!15}\xmark & \cellcolor{gray!15}W + S & \cellcolor{gray!15}\textbf{46.9} & \cellcolor{gray!15}\textbf{72.1} \\
            \midrule
            3 & 0.8 & center & \xmark & S & 45.2 & 51.8 \\
            3 & 0.8 & center & \xmark & W + S & 46.0 & 69.7 \\
            \bottomrule
        \end{tabular}}
        \caption{\textbf{Ablation of the number of self-training iterations in the proposed pipeline.} `W' denotes transcripts from WhisperX, and `S' denotes generated steps.}
        \label{table:ablations_iterations}
        
\end{table}



\section{Qualitative Results}

\subsection{ASR Transformation Example}

\begin{figure}[!t]
  \centering
   \includegraphics[width=.85\linewidth]{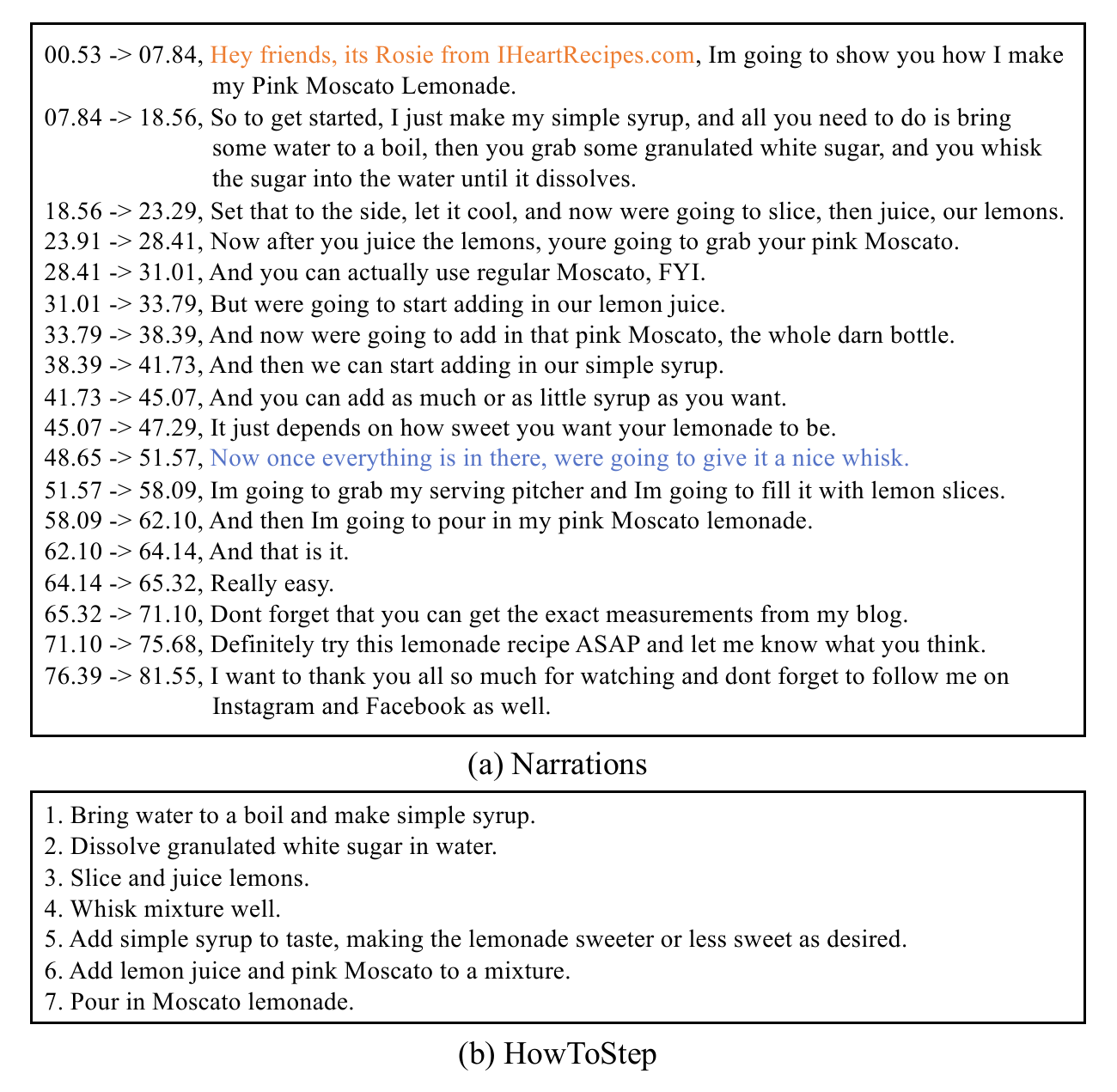}

   \caption{\textbf{A complete example of the ASR transformation.} (a) The ASR transcript recognized by WhisperX. (b) Descriptive steps transformed from the ASR transcript by the LLM. 
   }
   \label{fig:generate}
\end{figure}

In \cref{fig:generate}, we provide an example of using a LLM with the prompt to transform ASR transcripts into descriptive steps. 
Compared with descriptive steps, some complete accurately recognised ASR transcripts still contain semantic ambiguity and redundancy. 
For instance, in Fig. 2a, sentences in orange are irrelevant to the task at hand, while those highlighted in blue exhibit issues with unclear references.
Additionally, a segment of speech recognized by the ASR system often contains many important actions corresponding to different time intervals in the video, which significantly cause misalignment between text and video. Conversely, the descriptive steps generated by the LLM are more concise and clear, 
eliminating the redundant information in the ASR transcripts and articulating the action procedure in a unified form.

\subsection{Alignment Visualization}
\begin{figure*}[!t]
  \centering
   \includegraphics[width=\linewidth]{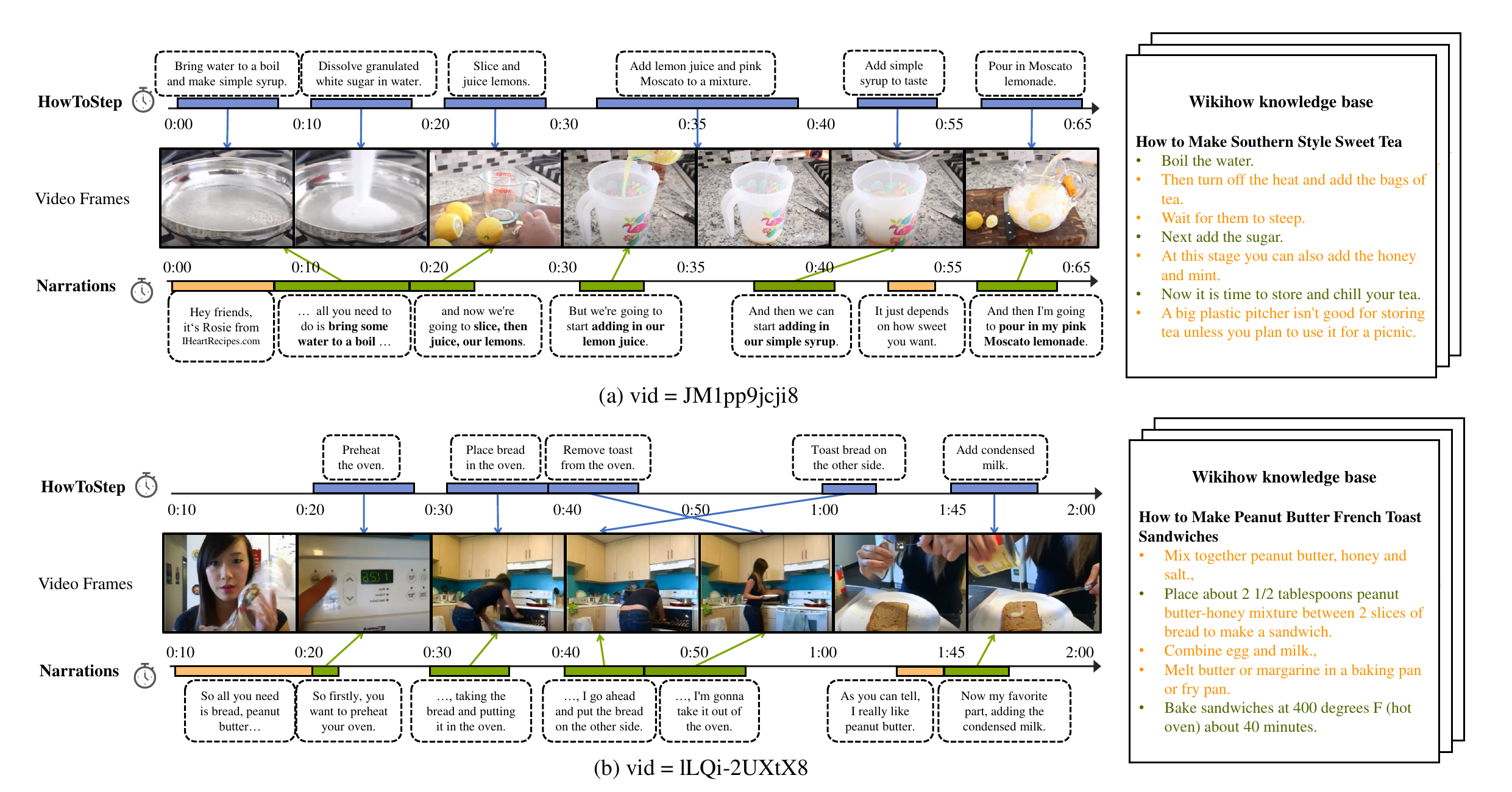}
   \caption{\textbf{Examples of alignment visualization.} \textit{(Left)}: The temporal distribution of descriptive steps in HowToStep and narrations in ASR transcripts, as well as their alignment with video frames. \textit{(Right)}: The procedural steps collected from an external knowledge base (Wikihow) to complete the task in the video without timestamps. For procedural steps in Wikihow, \textcolor{alignable}{\textbf{green}} timelines indicate the steps that can be aligned with the visual signals, while \textcolor{unalignable}{\textbf{yellow}} timelines represent steps that are not visually alignable.
   }
   \label{fig:visualization}
\end{figure*}

In \cref{fig:visualization}, we provide a visualization example of alignment for texts from various sources, including narrations from WhisperX transcripts, procedural steps from Wikihow, and descriptive steps in HowToStep. \cref{fig:visualization} clearly shows that narrations often suffer from severe misalignment with video content, while the steps in Wikihow are very generic and frequently inconsistent with the activities shown in the video, as color-coded with yellow.
However, our proposed HowToStep not only describes actions highly relevant to the video content using concise language but also demonstrates better temporal alignment after going through the two-stage temporal grounding process.
From \cref{fig:visualization}, it is evident that the majority of steps in the HowToStep are harmoniously aligned with the video content. However, the temporal misalignment of the sentences ``Remove toast from the oven.'' and ``Toast bread on the other side.'' in Fig. 3b arises due to obstructions, coupled with the challenge of distinguishing between fine-grained action such as taking out toast and flipping toast over.

\section{Limitations and Ethical Concerns}

As a proposal-free method for multi-sentence grounding, we do not explicitly generate the temporal window for each narration or procedural step, and only obtain it via post-processing the alignment score matrix. 
In addition, we pay more attention to whether the most possible timestamp for each text falls into the ground truth temporal window since the start-end boundary of the text is not used under the Recall@1 metric compared with Recall@IoU metrics. 
For ethical concerns, we are aware that the public instructional video dataset and the knowledge of large language models may have gender, age, geographical, or cultural bias.



\end{document}